\documentclass[a4paper, 10pt, conference]{ieeeconf}      

\IEEEoverridecommandlockouts                              

\usepackage{stfloats}
\usepackage{cite}
\usepackage{float}
\usepackage{bm}
\usepackage{here}
\usepackage{soul, color}
\usepackage{multirow}
\usepackage{tabularx}
\usepackage{algorithmic}
\usepackage{makecell}
\usepackage{hhline}
\usepackage{nicefrac}
\usepackage{lipsum}
\usepackage{array}
\usepackage{booktabs}
\usepackage{paralist}
\usepackage{gensymb}
\usepackage{inputenc}
\usepackage[dvipsnames]{xcolor}
\usepackage{colortbl}
\usepackage{svg}
\usepackage{threeparttable}
\usepackage{enumerate}
\usepackage{upgreek}
\usepackage[T1]{fontenc}
\usepackage{babel}
\usepackage{array}
\usepackage{textcomp}
\usepackage{mathtools}
\usepackage{amsmath}
\usepackage{amssymb}
\usepackage{graphicx}
\usepackage[unicode=true,
 bookmarks=true,bookmarksnumbered=true,bookmarksopen=true,bookmarksopenlevel=1,
 breaklinks=false,pdfborder={0 0 0},pdfborderstyle={},backref=false,colorlinks=false]
 {hyperref}
\hypersetup{pdftitle={Your Title},
 pdfauthor={Your Name},
 pdfpagelayout=OneColumn, pdfnewwindow=true, pdfstartview=XYZ, plainpages=false}

\usepackage[caption=false]{subfig}

\usepackage{babel}

\title{\LARGE \bf
Model-Based Control of Planar Piezoelectric Inchworm Soft Robot for
Crawling in Constrained Environments}
\author{Zhiwu Zheng, Prakhar Kumar, Yenan Chen, Hsin Cheng, \\Sigurd Wagner,
Minjie Chen, Naveen Verma and James C. Sturm\thanks{This work was supported by the Semiconductor Research Corporation
(SRC), DARPA, Princeton Program in Plasma Science and Technology,
and Princeton University. \emph{(Corresponding author: Zhiwu Zheng)}}\thanks{The authors are with the Department of Electrical and Computer Engineering,
Princeton University, Princeton, New Jersey 08544, U.S.A. (e-mail:
zhiwuz@princeton.edu; prakhark@princeton.edu; yenanc@princeton.edu;
hsin@princeton.edu; wagner@princeton.edu; minjie@princeton.edu; nverma@princeton.edu;
sturm@princeton.edu).}}
\begin{document}

\maketitle
\thispagestyle{empty}
\pagestyle{empty}

\begin{abstract}
Soft robots have drawn significant attention recently for their ability to achieve
rich shapes when interacting with complex environments. However, their elasticity 
and flexibility compared to rigid robots also pose significant challenges for
precise and robust shape control in real-time. Motivated by their potential to operate in 
highly-constrained environments, as in search-and-rescue operations, this work addresses
 these challenges of soft-robots by developing a model-based full-shape
controller, validated and demonstrated by experiments. A five-actuator planar soft robot was constructed with planar piezoelectric layers bonded to a steel foil substrate, enabling inchworm-like motion.
The controller uses a soft-body continuous
model for shape planning and control, given target shapes and/or environmental constraints, such as crawling under overhead barriers or "roof" safety lines. An approach to background model calibrations is developed to address
deviations of actual robot shape due to material parameter variations and drift. Full
experimental shape control and optimal movement under a roof safety line are demonstrated, where the robot maximizes its speed within the overhead constraint. The mean-squared error between the
measured and target shapes improves from \textasciitilde{} 0.05 cm$^{2}$ without calibration to \textasciitilde{} 0.01 cm$^{2}$
with calibration. Simulation-based validation is also performed with various different
roof shapes.
\end{abstract}

\section{Introduction}

Soft robots have gained attention due to their abilities to take on richer 
shapes than their traditional rigid counterparts. Particularly,
electrostatic soft robots, using piezoelectric actuators, enable flexible construction/integration, 
small form factors \cite{Jafferis2019}, and fast response times
\cite{Wu2019,Ji2019}. Further, a planar form factor enables piezoelectric-based 
soft robots to move within space-constrained environments, as expected in search-and-rescue applications, 
where a robot may be needed to squeeze into collapsed
construction gaps or around barriers from damage/debris. In this
work, we demonstrate inchworm-like robot crawling movement on a surface, under height constraints from the environment. 
We place an overhead barrier with various bottom
contours (referred to later as a \textquotedblleft roof\textquotedblright{}) on top of the robot, and require the robot to crawl with maximum speed while avoiding collision with the roof.

Precise and real-time shape control is needed for managing such environment constraints. For soft robots, this is challenging, as their shapes
are prominently affected by external forces, due to their elastic bodies.
Moreover, due to their flexible bodies, their shape  may possess many (infinite) degrees of freedom, requiring design decisions on how to feasibly model and process their state space.

\begin{figure}
\centering
\includegraphics[width=0.70\columnwidth]{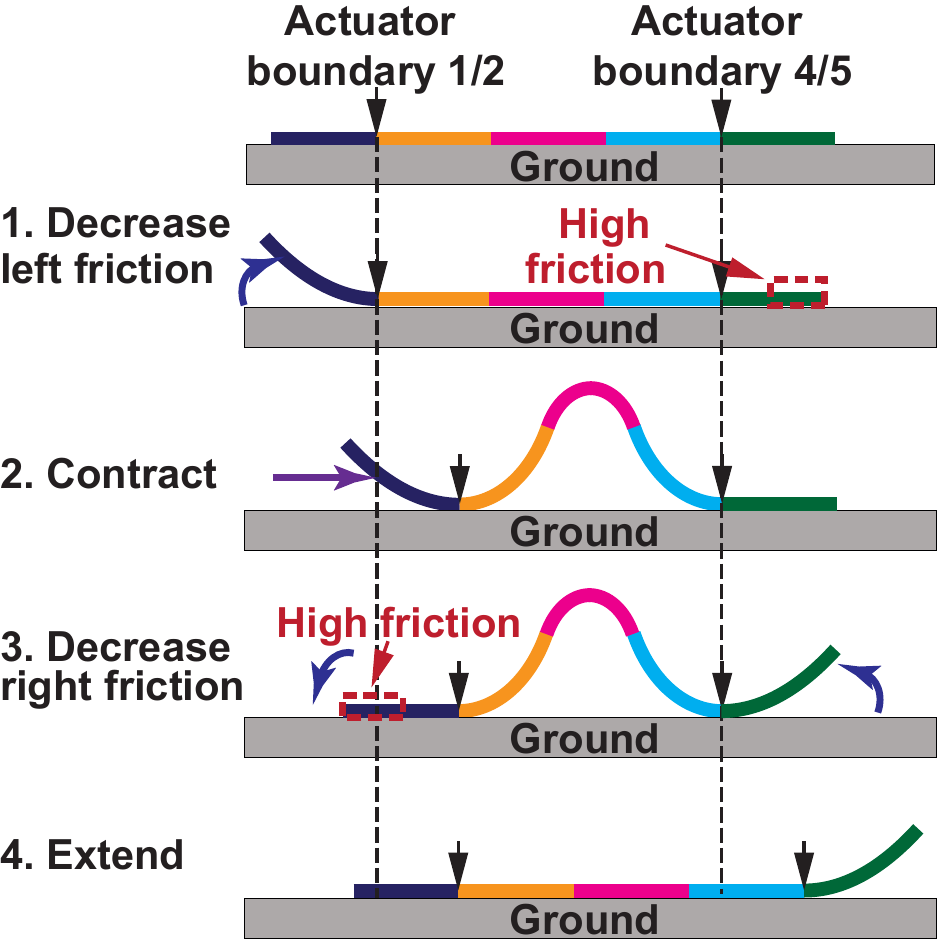}
\caption{Inchworm motion of the demonstrated robot, whereby the robot can crawl forward by straightening and bending while alternately lifting and lowering its high-friction ends. \label{fig:inchworm-overview}}
\end{figure}

This work focuses on model-based approaches to tackle these challenges.
Model-based control saves computational power and enables faster response speed.
Recent works have used the constant curvature model \cite{Webster2010,Falkenhahn2015,DellaSantina2020b}
or pseudo rigid body model \cite{Lobontiu2001,Bandopadhya2010,Li2018,Zheng2022}
for soft robots to discretize the infinite degrees of freedom (function spaces) into
finite degrees of freedom (n-d real number spaces). In this work, we propose and
develop for the first time a model-based full shape controller that
utilizes a soft and continuous body model developed in our previous work \cite{Zheng2021}
in function space for precise
and accurate control of electrostatic soft robots. Fig. \ref{fig:inchworm-overview} shows a five-actuator robot used
to demonstrate the controller. A four-phase inchworm-like motion is implemented,
by bending and straightening the robot's body, while lifting and lowering its ends which is highly-frictional undersides. The friction provides anchoring, and combined with bending and straightening leads to desired lateral motion.

This paper is organized as follows. Section \ref{sec:robot-structure}
introduces the robot structure and actuation mechanism. Section \ref{sec:model-based-shape-controller}
describes: (A) the proposed soft-body model-based shape controller;
(B) the controller for crawling under roofs with implicit shape planning;
and (C) background model calibration to address differences between
model-predicted and actual shapes, caused by material variations and
drift. Section \ref{sec:experimental-demonstrations} describes experimental
demonstrations of the controller for robot shape control and crawling
under a step-shaped roof.

\section{Robot structure\label{sec:robot-structure}}

\begin{figure}
\centering
\subfloat[\label{fig:robot-setup}]{\includegraphics[width=0.95\columnwidth]{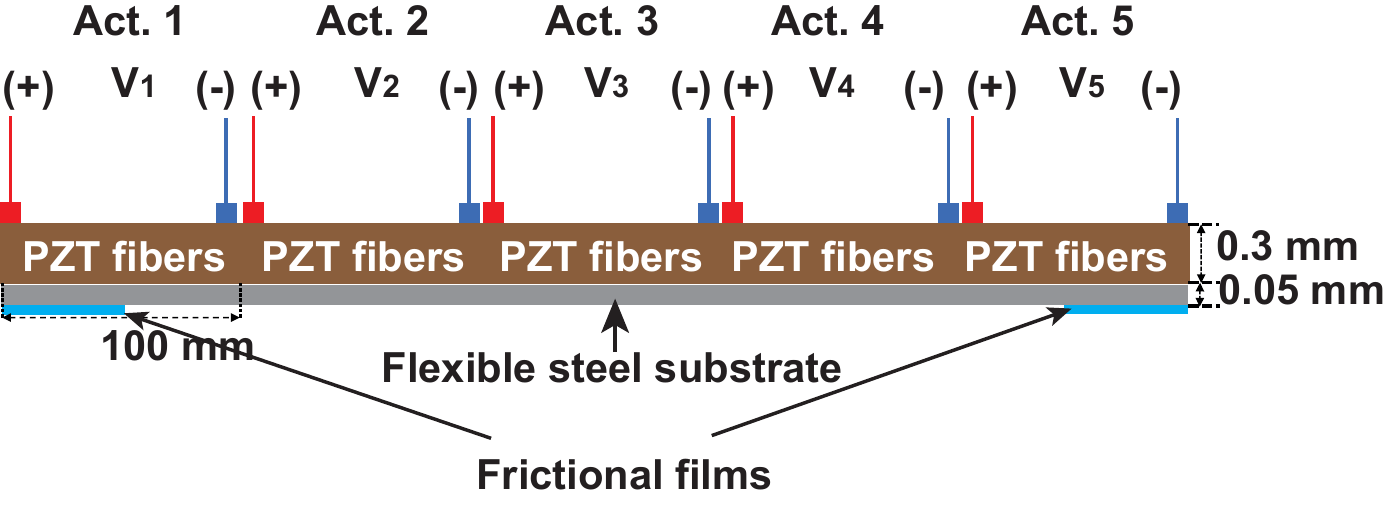}}

\subfloat[\label{fig:actuator-bending-mechanism}]{\includegraphics[width=0.95\columnwidth]{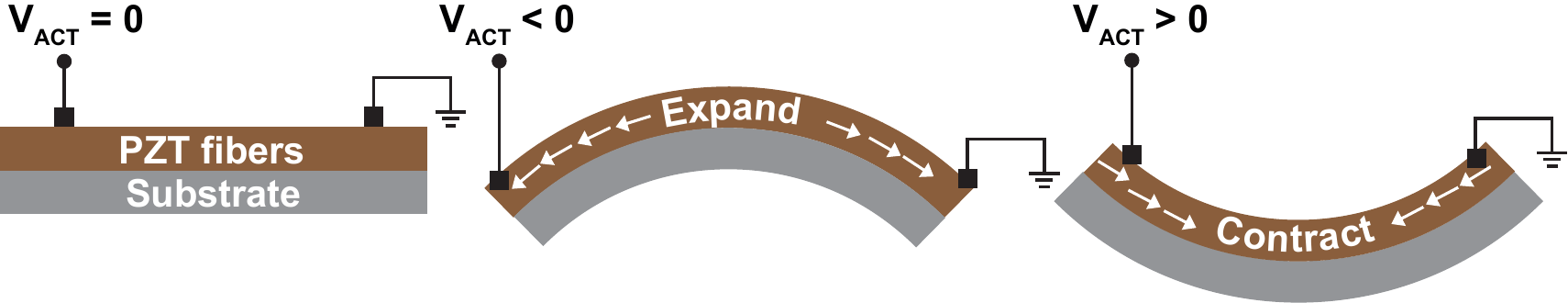}}

\subfloat[\label{fig:robot-top-picture}]{\includegraphics[width=0.95\columnwidth]{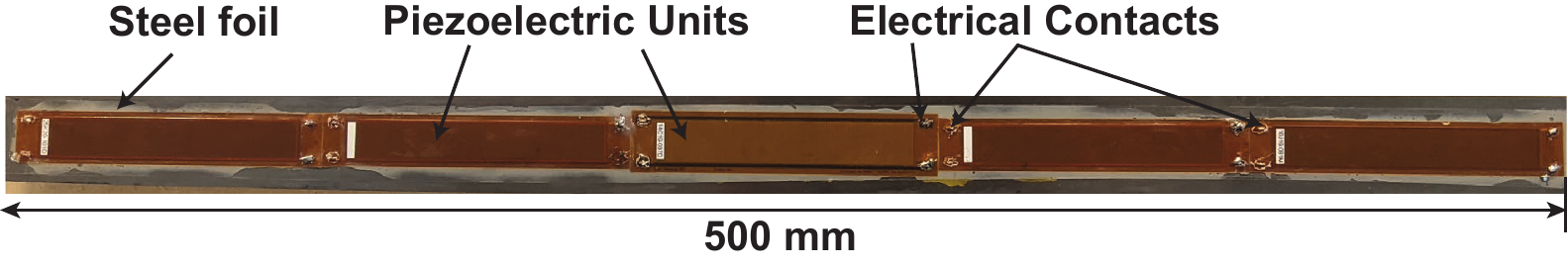}}

\caption{Structure of the experimental robot, showing: (a) cross-section of the five-actuator sections bonded to steel foil, with a high friction film applied to the extreme 50 mm ends on the underside;
(b) mechanism of bending for each unit actuator, via piezoelectric-composite expansion causing bending down with negative voltage and contraction causing bending up with positive voltage; (c) top view of the experimental robot prototype.}
\end{figure}

Fig. \ref{fig:robot-setup} shows the demonstrated robot structure.
The robot is 500 mm long and 20 mm wide. It contains five 100-mm-long 300-\textmu{}m-thick units of 
piezoelectric macrofiber composite bonded onto a 50-\textmu{}m-thick flexible steel foil. On the extreme left and right ends, high
friction films are bonded to its underside. During testing, the robot sits on a
flat surface and is driven by external voltage sources connected by
thin flexible wires.

Fig. \ref{fig:actuator-bending-mechanism} illustrates the bending
mechanism of a single actuator. When negative (positive) voltage is
applied to an actuator, the piezoelectric unit expands (contracts),
while the steel-foil substrate tends to remain stiff. As a result, the whole structure bends 
concave-down (-up). Fig. \ref{fig:robot-top-picture} shows a top view of the physical robot prototype,
including electrical contacts on the piezoelectric units.

\section{Model-based shape controller\label{sec:model-based-shape-controller}}

\begin{figure}
\centering
\subfloat[\label{fig:crawling-schematics}]{\includegraphics[width=0.70\columnwidth]{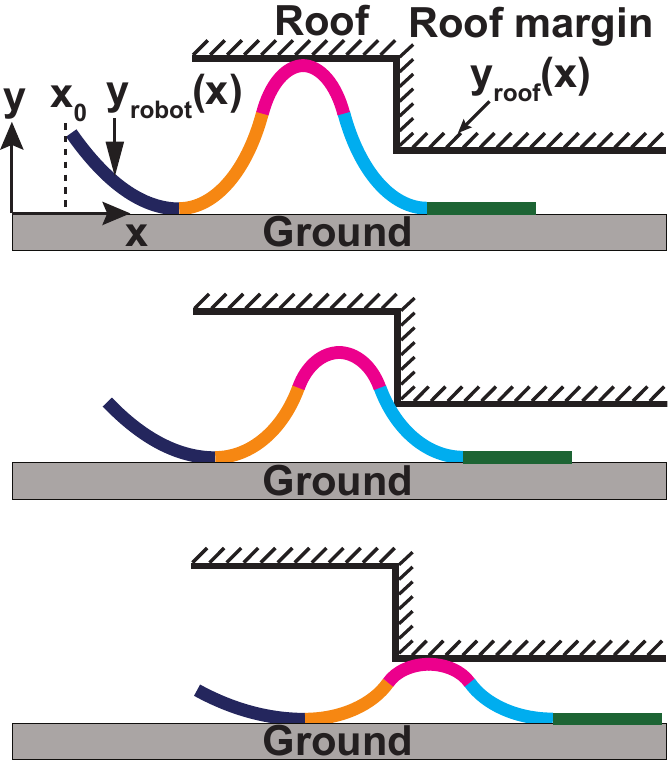}}

\subfloat[\label{fig:speed-bending-height}]{\includegraphics[width=0.70\columnwidth]{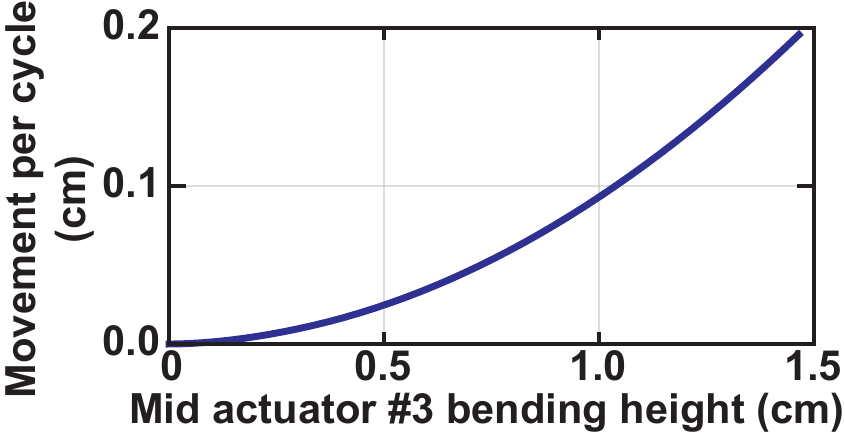}}

\caption{(a) Challenges of crawling under roofs, showing: (a) the critical point vulnerable
to collisions can be anywhere on the soft robot, necessitating
full-shape control; (b) the lateral movement per cycle of the inchworm robot with respect to bending height, predicted by the model, motivating the need to maximize height for maximum speed within the constraint of avoiding roof collisions.}
\end{figure}

For crawling under overhead barriers, full-shape control of the robot is needed.
Fig. \ref{fig:crawling-schematics} illustrates the process and challenges
of crawling under barriers with a step-shaped profile as an example.
A barrier called the ``roof'' is envisioned above the robot, creating
a height constraint during robot rightward movement. As shown in Fig. \ref{fig:speed-bending-height}, 
inchworm motion requires the midsection
of the robot to bend as high as possible for faster movement; but robot height must be controlled to avoid collisions
with the roof. When the robot is at the far left or right sections of the roof,
the critically vulnerable point for collision is at the midpoint of
the robot, where its height is maximum. 
However, when the robot is at the middle region, where the roof height changes sharply, the critical point can be anywhere on the robot profile. This
motivates full shape control and planning.

\begin{figure}
\centering
\includegraphics[width=1.0\columnwidth]{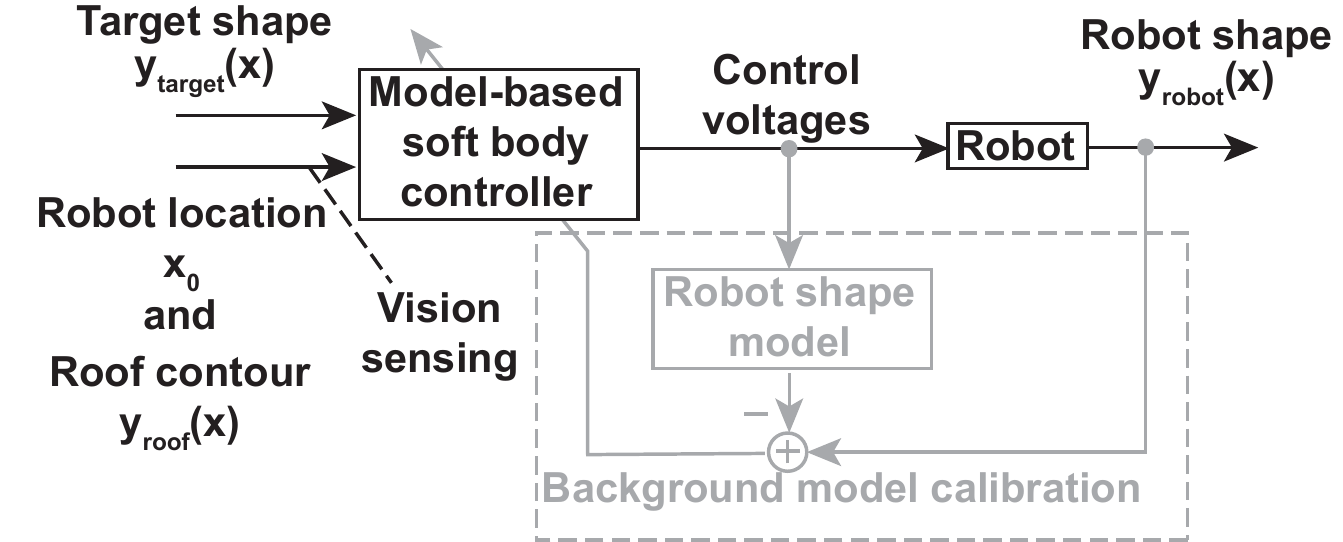}
\caption{System block diagram of the soft robot\label{fig:control-diagram}.}
\end{figure}

Fig. \ref{fig:control-diagram} shows a system block diagram of the robot, which includes a model-based soft-body controller and background calibration loop, in addition to the soft robot itself. For the controller,
a pre-specified target shape ($y_{\text{target}}(x)$) may be provided, or inputs from vision sensing, including the robot position ($x_0$) in an environment and the roof profile ($y_{\text{roof}}(x)$), may be provided. The controller then outputs control voltages to
the robot's five actuators. Background calibration of the model used by the controller can be run
before or continuously during operation. It uses vision sensing to compare the differences between model-predicted shapes and actual robot shapes, to determine the value of correction coefficients within the model.

Section \ref{sec:optimization-based-full-shape} discusses the model-based soft-body shape controller.
Section \ref{sec:controller-for-crawling} discusses the controller
for crawling under roofs with an implicit shape planner. Section
\ref{sec:background-model-calibration} discusses the background
model calibration.

\subsection{Optimization-based target shape controller \label{sec:optimization-based-full-shape}}

The goal of the target-shape controller is to control the robot to reach
a target shape $y_{\text{\text{target}}}(x)$ as closely as possible. This is represented by the following $L_{2}$-norm optimization objective:

\begin{equation}
\min_{V\in\mathbb{C}}||\hat{y}(x,V)-y_{\text{\text{target}}}(x)||,
\end{equation}
where $V$ is a vector of voltages that are applied to the actuators,
$\mathbb{C}$ represents all the possible voltages that can be applied, $\hat{y}(x,V)$ is the model-predicted robot soft-body shape (employing previous work on soft-robot modeling \cite{Zheng2021}). For our soft robots, gravity and ground forces affect the shape deformations. This soft-body-based model solves the deformed robot's shape due to gravity, piezoelectricity and ground interactions. The model is based on Euler-Bernoulli model, and its key property is a proposed self-consistent approach to determine which part of the robot is on the ground.

Using $L_{2}$ norm minimization, the optimization loss
function becomes: 

\begin{equation}
L(V)=||\hat{y}(x,V)-y_{\text{\text{target}}}(x)||_{2}^{2}=\int dx\left(\hat{y}(x,V)-y_{\text{\text{target}}}(x)\right)^{2},\label{eq:loss-function-shape-controller}
\end{equation}
whereby corresponding voltages $V$ and a model-predicted shape are thus selected from all possible shapes, that yield is closest to the target shape. Generally, $L(V)$ is expensive to evaluate. We employ Bayesian optimization
method \cite{Frazier2018} to solve this optimization problem, and validate the results experimentally, as described in Section \ref{sec:full-shape-control}. 

\subsection{Controller for crawling under a roof \label{sec:controller-for-crawling}}

Crawling under a roof represents a different but related task, compared to target shape
control. Rather than controlling for a pre-specified target shape, crawling under a roof requires shape planning to avoid collisions with the roof, while maximizing height for robot speed.  

\begin{figure}
\centering
\subfloat[\label{fig:crawling-setup-schematics-beyond}]{\includegraphics[width=0.70\columnwidth]{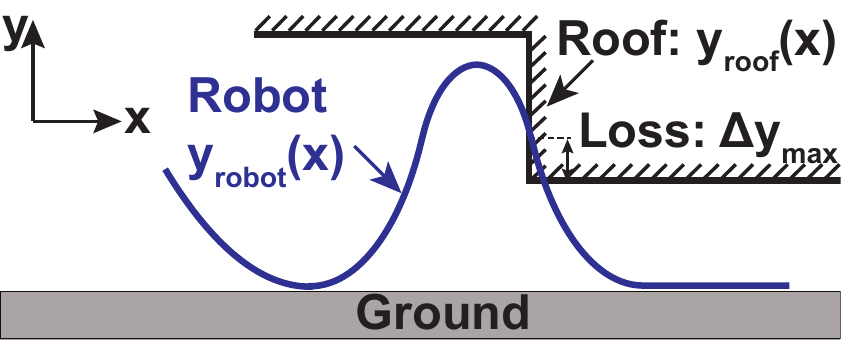}}

\subfloat[\label{fig:crawling-setup-schematics-below}]{\includegraphics[width=0.70\columnwidth]{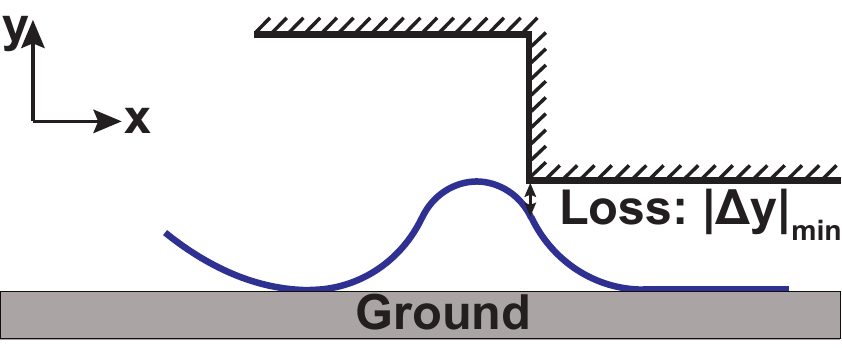}}

\subfloat[\label{fig:crawling-setup-schematics-optimal}]{\includegraphics[width=0.70\columnwidth]{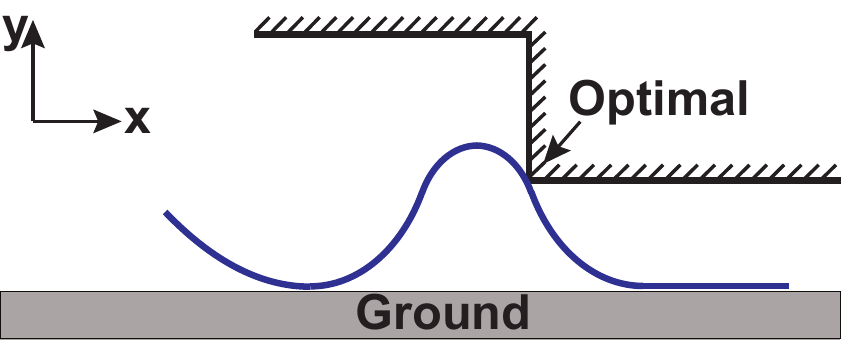}}

\caption{Illustration of robot shape-planning objective for crawling under a roof with maximum speed, showing three cases: (a) robot going beyond the roof safety line, which is not allowed; (b) robot below the roof, which leads to reduced speed; and (c)
robot touching the roof safety line, which leads to maximum speed without roof collision. \label{fig:crawling-setup-schematics}}
\end{figure}

Fig. \ref{fig:crawling-setup-schematics} illustrates three cases
during crawling:
\begin{enumerate}
\item The robot bends too much, such that its shape ($y_{\text{robot}}(x)$)
goes beyond the roof safety line ($y_{\text{roof}}(x)$), as shown in (Fig. \ref{fig:crawling-setup-schematics-beyond}).
This implies that the robot would collide with the roof, and is not desired.
\item The robot bends too little, such that its shape is below the roof safety line, as shown in
(Fig. \ref{fig:crawling-setup-schematics-below}). This avoids roof collision, but leads to reduced speed, and is thus also not desired.
\item The robot bends so as to just ``touch'' the roof safety line, as shown in Fig. \ref{fig:crawling-setup-schematics-optimal}.
In this case, collision is avoided and the robot bends as much as
possible, achieving maximum speed,
\end{enumerate}

We implement the controller for this task by changing the loss function
to be:

\begin{equation}
L=\begin{cases}
\Delta y_{\text{max}}+c & \text{if }\Delta y_{\text{max}}>0\\
|\Delta y|_{\text{min}} & \text{elsewhere},
\end{cases}
\end{equation}
where $c>0$ is an arbitrarily large positive constant to heavily penalize
collisions, and:

\begin{equation}
\Delta y_{\text{max}}=\max_{x}\left(y_{\text{robot}}(x)-y_{\text{roof}}(x)\right)
\end{equation}

\begin{equation}
|\Delta y|_{\text{min}}=\min_{x}|y_{\text{robot}}(x)-y_{\text{roof}}(x)|.
\end{equation}

For this loss function, if the robot is in case (1), the loss will
take on a large positive number, prioritizing avoiding collisions.
If the robot is in case (2), the loss value will also be a positive
number, pushing towards reducing the separation to the roof safety line. However, if the robot is in case (3), the loss is zero. Therefore, minimizing this loss function results in optimal crawling motion to be achieved.
This resulting controller is experimentally validated and demonstrated, as discussed
later in Section \ref{subsec:crawling-under-roof-demonstration}.

\subsection{Background model calibration \label{sec:background-model-calibration}}

Generally, the real robot shape will deviate from model-predicted shape. This can
be due to materials variations (of the piezoelectric-composite actuators, steel foil, bonding adhesive), fabrication/assembly variations (adhesive thickness), and drift over time (elasticity of the components, charge fluctuations in the piezoelectrics). While this deviation may be small, but it can easily exceed to margins available in constrained scenarios, such as crawling under roof safety lines to avoid hazardous collisions. Thus, background model calibration is employed.

The calibration approach augments the physics-based shape model (developed in the previous work \cite{Zheng2021}) with a linear correction term to enable compensation of shape differences. We replace $\hat{y}(x,V)$ with the original model-predicted shape $\tilde{y}(x,V)$
plus a correction term $\Delta y(x,V)$:

\begin{equation}
\hat{y}(x,V)=\tilde{y}(x,V)+\Delta y(x,V)
\end{equation}

Since $\Delta y(x,V)$ is small, we restrict to a linear approximation, with respect to the actuator input voltages, for
this correction term, as follows:

\begin{equation}
\Delta y(x,V)\sim\alpha(x)^{T}V
\end{equation}
where $\alpha(x)$ is a vector as a function of $x$, representing the proportional shape correction.

Background calibration can be performed prior to the robot operation
to deal with time-invariant deviations or during usage of the robot to deal with drifting
deviations over time. For demonstration, in this work we only perform
prior calibration. 

During calibration, $\Delta y(x,V)$ is the difference between experimentally observed robot shape through vision sensing $y_{\text{robot}}(x)$ and the model-predicted shape $\tilde{y}(x,V)$. For each $x$, $\alpha(x)$
is updated using online linear regression \cite{Ruano2005}:

\begin{equation}
\alpha(x)\gets\alpha(x)-\eta\left[\alpha(x)^{T}V-\Delta y(x,V)\right]V
\end{equation}
where $\eta>0$ is the learning rate, and initially $\alpha(x)=0$.

\section{Experimental demonstrations\label{sec:experimental-demonstrations}}

\begin{figure}
\centering
\includegraphics[width=0.80\columnwidth]{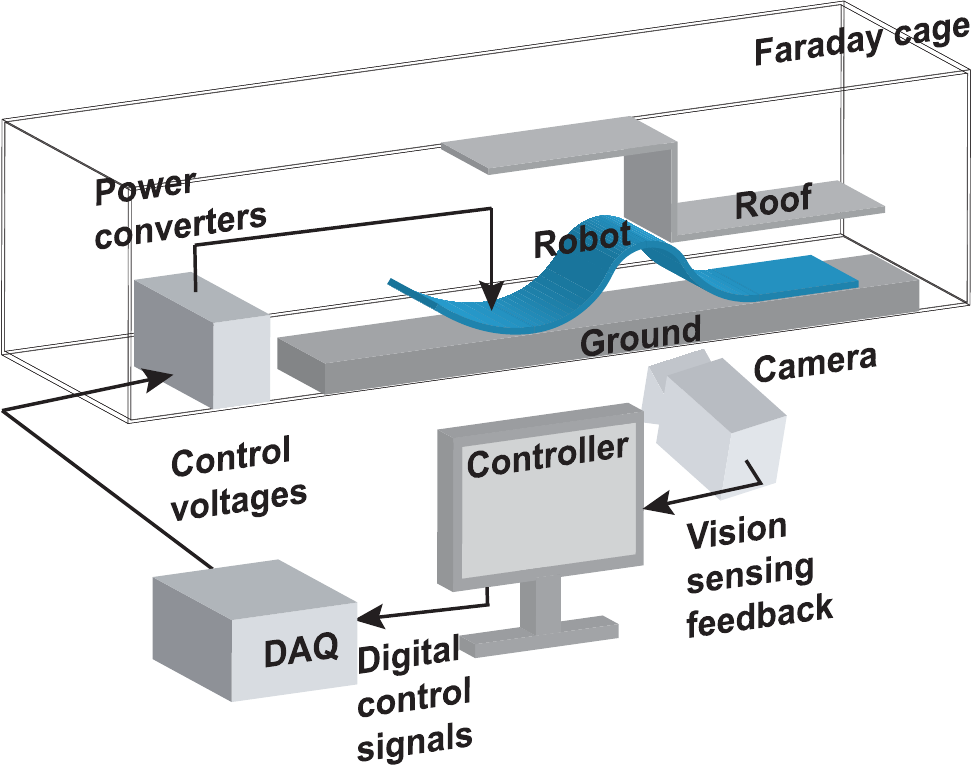}

\caption{Experimental setup block diagram for the experimental demonstrations. \label{fig:3d-test-setup}}
\end{figure}

Fig. \ref{fig:3d-test-setup} shows experimental setup for the demonstrations. The robot is placed in a Faraday cage, and commercial power converters drive the actuators. The power converters are controlled by a data acquisition system (DAQ). Camera-based vision sensing can track the shape and location of the robot in real-time using edge detection. The controller uses the vision sensing feedback to determine the actuator voltages through the DAQ.

\subsection{Target-shape control demonstration\label{sec:full-shape-control}}

\begin{figure}
\centering
\includegraphics[width=0.70\columnwidth]{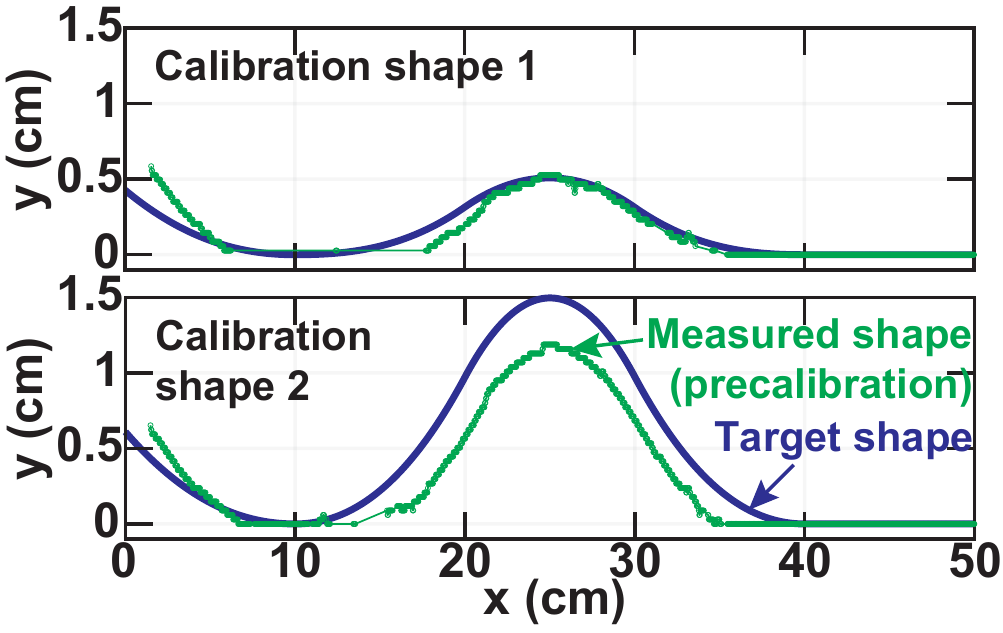}

\caption{Background model calibration employing two reference shapes measured from vision sensing and compared against the model-predicted shapes. \label{fig:shapes-for-calibration}}
\end{figure}

Prior model calibration is performed using two reference shapes, by applying the actuator voltages and measuring the resulting robot shapes from vision sensing, for comparison with model-predicted shapes. Fig. \ref{fig:shapes-for-calibration}
shows these measured shapes and the model-predicted shapes used for calibration. In the first case, the model-predicted shape achieves relatively good match with the measured shape. In the second case, the deviation is considerable. The deviations in both cases are used for calibration. 

\begin{figure}
\centering
\subfloat[\label{fig:shape-controller-shape}]{\includegraphics[width=0.70\columnwidth]{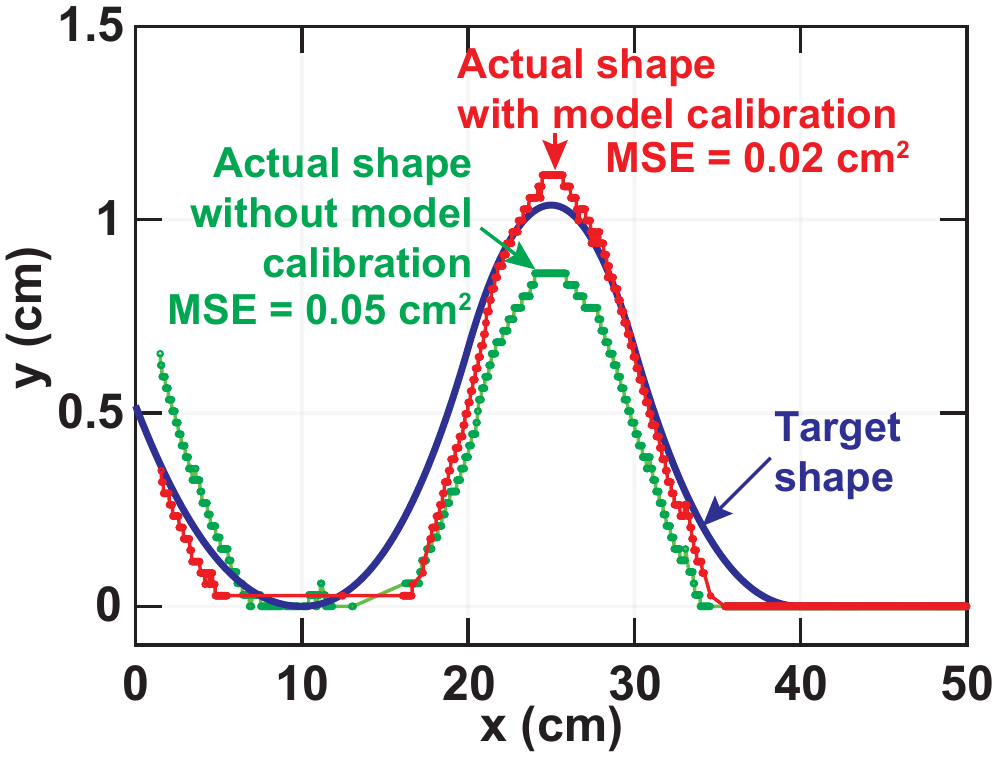}

}

\subfloat[\label{fig:shape-controller-height}]{\includegraphics[width=0.70\columnwidth]{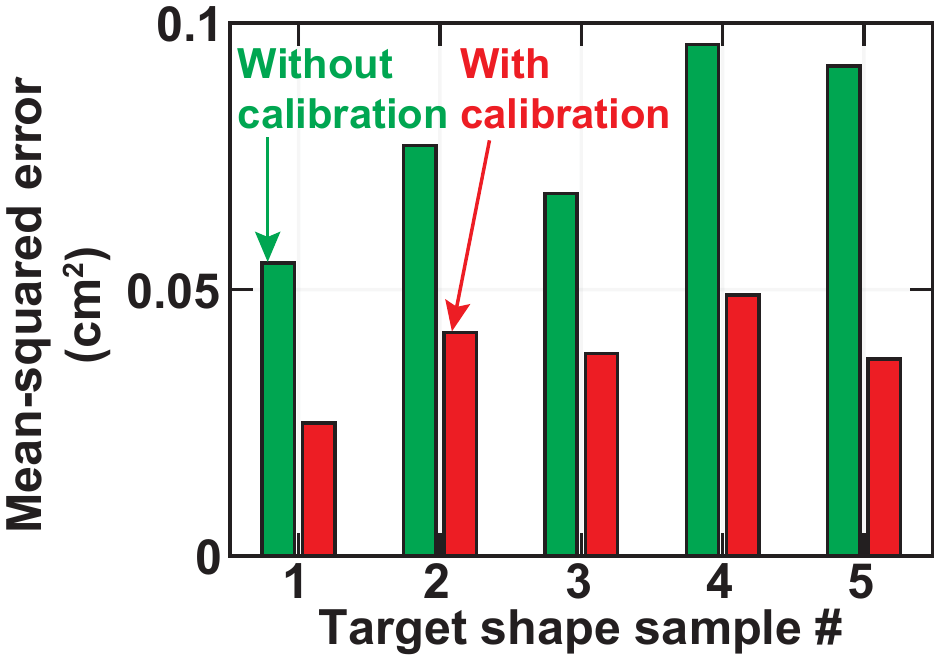}

}

\caption{Shape controller validations for a five-actuator robot, showing: (a) shape-control 
results for a particular target shape, with pre/post-calibrated measured shapes overlaid; (b) mean-squared errors
for five different target shapes both before calibration (green) and after calibration (red).}
\end{figure}

After calibration, the controller starts to perform shape-control
tasks. Fig. \ref{fig:shape-controller-shape} shows an example test
target shape that the robot is controlled towards. The control results
for models without and with calibrations are overlaid with the target
shape. The measured shape after calibration shows much better match with the target shape, with mean-squared error
(MSE) of $0.02$ $\text{cm}^{2}$. While the measured shape before calibration has an MSE of $0.05\text{ cm}^{2}$.

Fig. \ref{fig:shape-controller-height} compares
the pre- and post-calibration MSEs for five different target shapes (corresponding to different inchworm phases). Relatively small deviations are observed in all cases, with calibration further improving the match by roughly 50\%. The residual error is thought to result from non-idealities in the robot causing its shape to differ from that predicted by the physics-based model.

\subsection{Crawling under roofs demonstration\label{subsec:crawling-under-roof-demonstration}}

\begin{figure}
\centering
\subfloat[Voltages = (300, 258, -1292, 258, 300) volts.\label{fig:robot-setup-pictures-left}]{\includegraphics[width=\columnwidth]{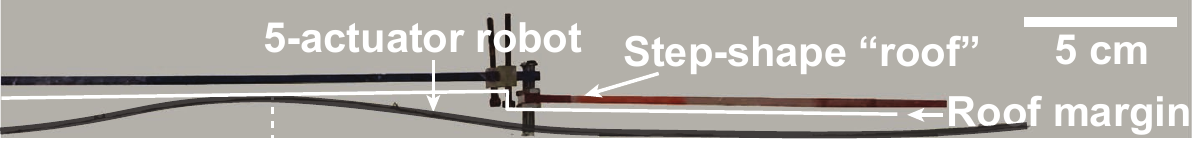}
}

\subfloat[Voltages = (300, 232, -1160, 232, 300) volts.\label{fig:robot-setup-pictures-mid}]{\includegraphics[width=\columnwidth]{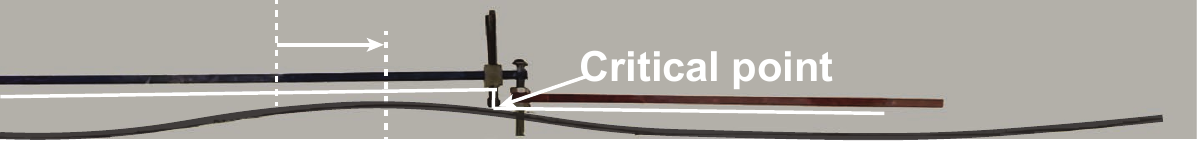}
}

\subfloat[Voltages = (300, 175, -800, 175, 300) volts.\label{fig:robot-setup-pictures-right}]{\includegraphics[width=\columnwidth]{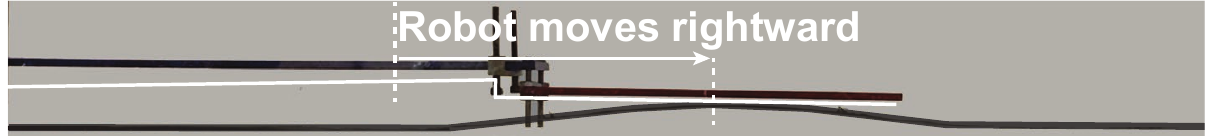}
}
\caption{Robot side view when crawling under a step-shape roof (with 1 mm margin to the roof explicitly added, as the white dashed lines), showing:   (a) crawling under the left (high) part; (b) the interface and finally (c)
the right part during crawling. \label{fig:robot-setup-pictures}}
\end{figure}

Fig. \ref{fig:robot-setup-pictures} shows experimental snapshots
of the robot crawling on an acrylic board, while crawling under a step-shaped roof.
The roof is placed on top of the robot as an overhead barrier, and the robot crawls and moves to the
right, as the controller evaluates actuator voltages in real-time. The roof safety line and relative robot position are determined from vision sensing, and a margin of 1 mm from the roof is explicitly added (to validate robot height achieved is as controlled, and not due to collision with the roof). The snapshots show the robot at the high (left) section (Fig. \ref{fig:robot-setup-pictures-left}), the transition section (Fig. \ref{fig:robot-setup-pictures-mid}), and finally
to the low (right) section (Fig. \ref{fig:robot-setup-pictures-right}).
The dashed line under the roof shows the pre-specified margin.
Note, when the robot is at the transition section, the critical point is not its maximum-height midpoint, thus necessitating full-shape planning and control as it passes through.

\begin{figure}
\centering
\subfloat[\label{fig:crawling-demonstration}]{\includegraphics[width=0.75\columnwidth]{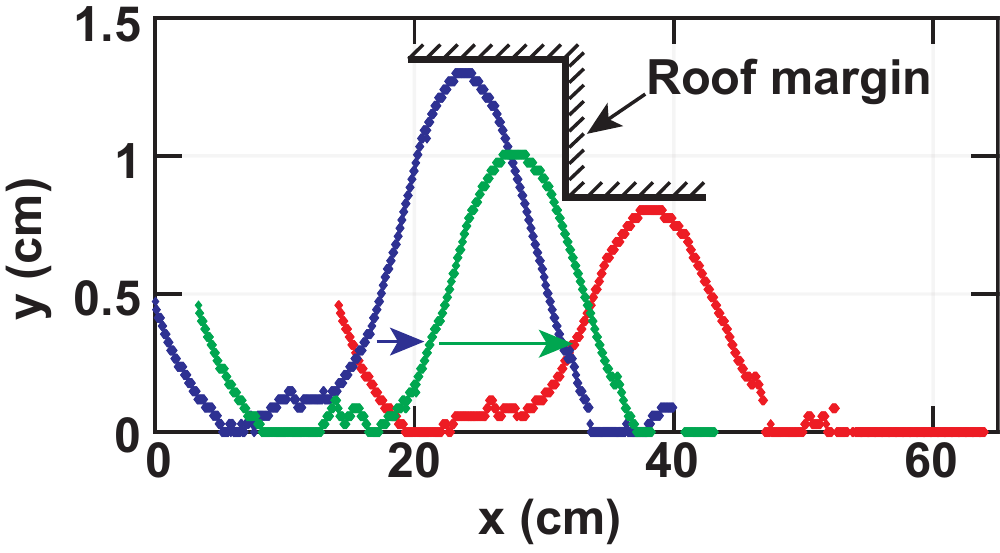}

}

\subfloat[\label{fig:speed-robot-position}]{\includegraphics[width=0.65\columnwidth]{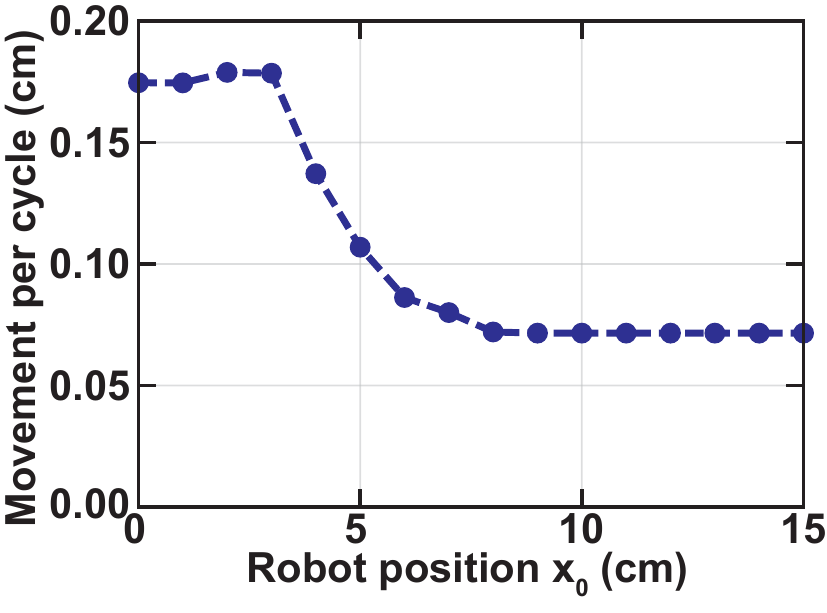}

}

\caption{(a) Experimental demonstration of crawling under a roof: robot shapes
for different positions during the crawling; (b) model-predicted robot
movement speed as a function of robot position.}
\end{figure}

Fig. \ref{fig:crawling-demonstration} quantitatively shows robot
shapes (blue, green, and red curves) at the three positions during
crawling (pictures shown in Fig. \ref{fig:robot-setup-pictures}).
The robot crawls through a step-shaped roof from 1.4 cm
high to 0.9 cm. The robot remains below the roof safety line (which has the specified roof margin from the roof). Robot
shapes at the three different positions are shown, and their closest distances
to the roof safety line are 0.04, 0.06, 0.04 cm, respectively.

Fig. \ref{fig:speed-robot-position} shows the model-predicted robot
movement speed, represented by movement distance per driving cycle,
as a function of the robot position. The robot's speed is $\sim0.17$
cm/cycle at the high (left) part. It decreases when the robot reaches
the transition section, and reaches $\sim0.07$ cm/cycle when the robot is
at the (low) right part. This shows that the robot maintains a compromise between speed and avoiding roof collisions during the course of its movement.

\begin{figure}
\centering
\subfloat[\label{fig:crawling-simulation-linear}]{\includegraphics[width=0.60\columnwidth]{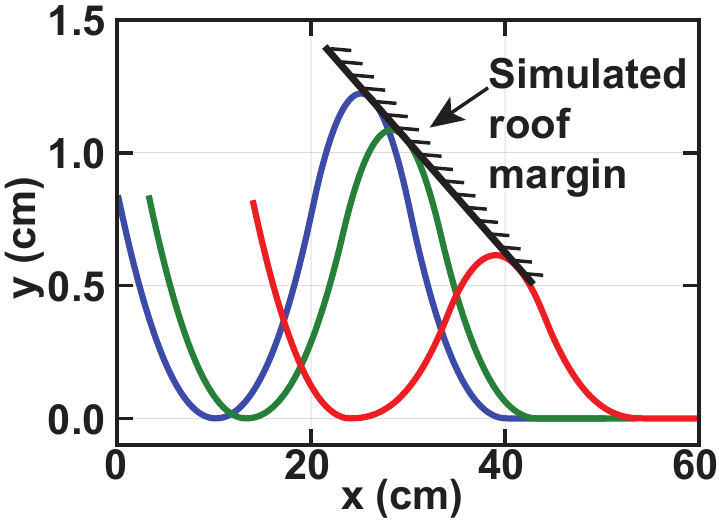}

}

\subfloat[\label{fig:crawling-simulation-sinusoidal}]{\includegraphics[width=0.60\columnwidth]{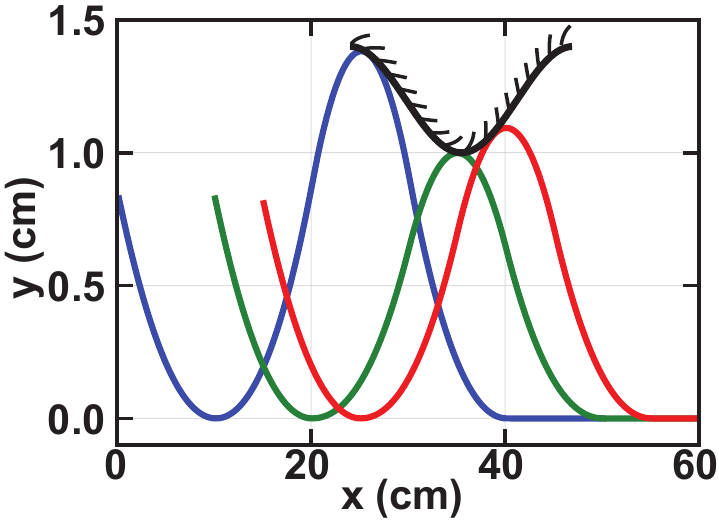}

}

\caption{Simulation validation of the controller working with two different roof
shapes: (a) slanted and (b) sinusoidal. \label{fig:crawling-simulation}}
\end{figure}

Fig. \ref{fig:crawling-simulation} shows validation of the controller
by simulations for two different roof shapes: slanted (Fig.
\ref{fig:crawling-simulation-linear}) and sinusoidal (Fig. \ref{fig:crawling-simulation-sinusoidal}).
Robot shapes at different stages of crawling are shown (blue, green,
and red curves), illustrating that the robot maintains optimal shape for different
roof shapes.

\section{Conclusion}

The development of soft robot controllers is driven by the their ability to take various shapes and motions, for movement within complex environments. This work proposes a model-based
soft robot shape controller demonstrated with a five-actuator piezoelectric
planar soft robot, performing inchworm-like crawling under roofs with
height constraints. The controller is based on optimization for the
differences between specified target shapes and model-predicted shapes.
Furthermore, for crawling under roofs, a loss function is proposed
for implicit shape planning, based on the distance between the robot
and the roof. In addition, background model calibration is implemented
for improving control against material variations and drifts.
The controller is validated by experiments demonstrating target-shape control and
crawling under a step-shaped roof. Mean-squared errors
for shape control are $\sim0.05\text{ cm}^{2}$ before calibration,
and reduced to $\sim0.01\text{ cm}^{2}$ after calibration, using
two reference shapes. Moreover, for the crawling demonstration,
$<1\text{ mm}$ distances between the pre-specified roof safety line and the robot are achieved.
The controller is also validated for different roof shapes by simulations. The controller
will be further used in more complex settings and interactions between the robot and its
environment.

\bibliographystyle{IEEEtran}
\bibliography{IEEEabrv, mybib}

\begin{thebibliography}{10}
\providecommand{\url}[1]{#1}
\csname url@rmstyle\endcsname
\providecommand{\newblock}{\relax}
\providecommand{\bibinfo}[2]{#2}
\providecommand\BIBentrySTDinterwordspacing{\spaceskip=0pt\relax}
\providecommand\BIBentryALTinterwordstretchfactor{4}
\providecommand\BIBentryALTinterwordspacing{\spaceskip=\fontdimen2\font plus
\BIBentryALTinterwordstretchfactor\fontdimen3\font minus
  \fontdimen4\font\relax}
\providecommand\BIBforeignlanguage[2]{{%
\expandafter\ifx\csname l@#1\endcsname\relax
\typeout{** WARNING: IEEEtran.bst: No hyphenation pattern has been}%
\typeout{** loaded for the language `#1'. Using the pattern for}%
\typeout{** the default language instead.}%
\else
\language=\csname l@#1\endcsname
\fi
#2}}

\bibitem{Jafferis2019}
\BIBentryALTinterwordspacing
N.~T. Jafferis, E.~F. Helbling, M.~Karpelson, and R.~J. Wood, ``{Untethered
  flight of an insect-sized flapping-wing microscale aerial vehicle},''
  \emph{Nature}, vol. 570, no. 7762, pp. 491--495, jun 2019. [Online].
  Available: \url{http://www.nature.com/articles/s41586-019-1322-0}
\BIBentrySTDinterwordspacing

\bibitem{Wu2019}
\BIBentryALTinterwordspacing
Y.~Wu, J.~K. Yim, J.~Liang, Z.~Shao, M.~Qi, J.~Zhong, Z.~Luo, X.~Yan, M.~Zhang,
  X.~Wang, R.~S. Fearing, R.~J. Full, and L.~Lin, ``{Insect-scale fast moving
  and ultrarobust soft robot},'' \emph{Science Robotics}, vol.~4, no.~32, p.
  eaax1594, jul 2019. [Online]. Available:
  \url{https://robotics.sciencemag.org/lookup/doi/10.1126/scirobotics.aax1594}
\BIBentrySTDinterwordspacing

\bibitem{Ji2019}
\BIBentryALTinterwordspacing
X.~Ji, X.~Liu, V.~Cacucciolo, M.~Imboden, Y.~Civet, A.~{El Haitami}, S.~Cantin,
  Y.~Perriard, and H.~Shea, ``{An autonomous untethered fast soft robotic
  insect driven by low-voltage dielectric elastomer actuators},'' \emph{Science
  Robotics}, vol.~4, no.~37, p. eaaz6451, dec 2019. [Online]. Available:
  \url{https://robotics.sciencemag.org/lookup/doi/10.1126/scirobotics.aaz6451}
\BIBentrySTDinterwordspacing

\bibitem{Webster2010}
\BIBentryALTinterwordspacing
R.~J. Webster and B.~A. Jones, ``{Design and Kinematic Modeling of Constant
  Curvature Continuum Robots: A Review},'' \emph{The International Journal of
  Robotics Research}, vol.~29, no.~13, pp. 1661--1683, nov 2010. [Online].
  Available: \url{http://journals.sagepub.com/doi/10.1177/0278364910368147}
\BIBentrySTDinterwordspacing

\bibitem{Falkenhahn2015}
\BIBentryALTinterwordspacing
V.~Falkenhahn, A.~Hildebrandt, R.~Neumann, and O.~Sawodny, ``{Model-based
  feedforward position control of constant curvature continuum robots using
  feedback linearization},'' in \emph{Proceedings - IEEE International
  Conference on Robotics and Automation}, vol. 2015-June, no. June.\hskip 1em
  plus 0.5em minus 0.4em\relax IEEE, may 2015, pp. 762--767. [Online].
  Available: \url{http://ieeexplore.ieee.org/document/7139264/}
\BIBentrySTDinterwordspacing

\bibitem{DellaSantina2020b}
\BIBentryALTinterwordspacing
C.~{Della Santina}, A.~Bicchi, and D.~Rus, ``{On an Improved State
  Parametrization for Soft Robots With Piecewise Constant Curvature and Its Use
  in Model Based Control},'' \emph{IEEE Robotics and Automation Letters},
  vol.~5, no.~2, pp. 1001--1008, apr 2020. [Online]. Available:
  \url{https://ieeexplore.ieee.org/document/8961972/}
\BIBentrySTDinterwordspacing

\bibitem{Lobontiu2001}
\BIBentryALTinterwordspacing
N.~Lobontiu, M.~Goldfarb, and E.~Garcia, ``{A piezoelectric-driven inchworm
  locomotion device},'' \emph{Mechanism and Machine Theory}, vol.~36, no.~4,
  pp. 425--443, apr 2001. [Online]. Available:
  \url{https://linkinghub.elsevier.com/retrieve/pii/S0094114X00000562}
\BIBentrySTDinterwordspacing

\bibitem{Bandopadhya2010}
\BIBentryALTinterwordspacing
D.~Bandopadhya, ``{Derivation of Transfer Function of an IPMC Actuator Based on
  Pseudo-Rigid Body Model},'' \emph{Journal of Reinforced Plastics and
  Composites}, vol.~29, no.~3, pp. 372--390, feb 2010. [Online]. Available:
  \url{http://journals.sagepub.com/doi/10.1177/0731684408097778}
\BIBentrySTDinterwordspacing

\bibitem{Li2018}
\BIBentryALTinterwordspacing
W.-B. Li, W.-M. Zhang, H.-X. Zou, Z.-K. Peng, and G.~Meng, ``{A Fast Rolling
  Soft Robot Driven by Dielectric Elastomer},'' \emph{IEEE/ASME Transactions on
  Mechatronics}, vol.~23, no.~4, pp. 1630--1640, aug 2018. [Online]. Available:
  \url{https://ieeexplore.ieee.org/document/8365835/}
\BIBentrySTDinterwordspacing

\bibitem{Zheng2022}
\BIBentryALTinterwordspacing
Z.~Zheng, P.~Kumar, Y.~Chen, H.~Cheng, S.~Wagner, M.~Chen, N.~Verma, and J.~C.
  Sturm, ``{Scalable Simulation and Demonstration of Jumping Piezoelectric 2-D
  Soft Robots},'' feb 2022. [Online]. Available:
  \url{http://arxiv.org/abs/2202.13521}
\BIBentrySTDinterwordspacing

\bibitem{Zheng2021}
\BIBentryALTinterwordspacing
------, ``{Piezoelectric Soft Robot Inchworm Motion by Controlling Ground
  Friction through Robot Shape},'' nov 2021. [Online]. Available:
  \url{http://arxiv.org/abs/2111.00944}
\BIBentrySTDinterwordspacing

\bibitem{Frazier2018}
\BIBentryALTinterwordspacing
P.~I. Frazier, ``{A Tutorial on Bayesian Optimization},'' no. Section 5, pp.
  1--22, 2018. [Online]. Available: \url{http://arxiv.org/abs/1807.02811}
\BIBentrySTDinterwordspacing

\bibitem{Ruano2005}
A.~E. Ruano, \emph{{Intelligent control systems using computational
  intelligence techniques}}.\hskip 1em plus 0.5em minus 0.4em\relax Institution
  of Electrical Engineers, 2005.

\end{thebibliography}

\end{document}